\documentclass[letterpaper]{article} 
\usepackage{aaai24}  
\usepackage{times}  
\usepackage{helvet}  
\usepackage{courier}  
\usepackage[hyphens]{url}  
\usepackage{graphicx} 
\usepackage{natbib}  
\usepackage{caption} 
\usepackage{latexsym}
\usepackage[dvipsnames]{xcolor}

\usepackage{tcolorbox}

\usepackage[utf8]{inputenc}

\usepackage{microtype}

\usepackage{inconsolata}

\usepackage{comment}
\usepackage{amsmath,amsthm,amssymb,bm}
\usepackage{cleveref}
\usepackage{booktabs}
\usepackage{multirow}

\usepackage{amsfonts}
\usepackage{caption}
\usepackage{subcaption}
\usepackage{makecell}

\usepackage{algorithm}
\usepackage{algorithmic}

\usepackage{tikz}
\usetikzlibrary{decorations.pathreplacing,calligraphy}

\usepackage{newfloat}
\usepackage{listings}
\DeclareCaptionStyle{ruled}{labelfont=normalfont,labelsep=colon,strut=off} 
\floatstyle{ruled}
\newfloat{listing}{tb}{lst}{}
\floatname{listing}{Listing}
\DeclareMathOperator*{\argmax}{argmax}

\def\aopck{\mathsf{AOPC}^k}
\def\gb{\mathcal{G}_{\mathsf{base}}}
\def\grev{\mathcal{G}^{-1}_{\mathsf{pr}}}
\def\gavemod{\mathcal{G}^{+5}_{\mathsf{ave}}}

\def\qmin{q_{\mathsf{min}}}
\def\wmin{w_{\mathsf{min}}}

\title{Accelerating the Global Aggregation of Local Explanations}
\author {
    Alon Mor, 
    Yonatan Belinkov, 
    Benny Kimelfeld  \\ 
}

\affiliations {
    Technion -- Israel Institute of Technology, Haifa, Israel \\ 
    almr16@campus.technion.ac.il \hspace{0.5em} belinkov@technion.ac.il \hspace{0.5em} bennyk@cs.technion.ac.il
}

\newcommand{\eat}[1]{}
\def\e#1{\emph{#1}}
  
\begin{document}
\maketitle
\begin{abstract}
Local explanation methods highlight the input tokens that have a considerable impact on the outcome of classifying the document at hand. For example, the Anchor algorithm applies a statistical analysis of the sensitivity of the classifier to changes in the tokens. Aggregating local explanations over a dataset provides a global explanation of the model.
Such aggregation aims to detect words with the most impact, giving valuable insights about the model, like what it has learned in training and which adversarial examples expose its weaknesses.
However, standard aggregation methods bear a high computational cost:
a na\"ive implementation applies a costly algorithm to each token of each document, and hence, it is infeasible for a simple user running in the scope of a short analysis session.  

We devise techniques for accelerating the global aggregation of the Anchor algorithm. Specifically, our goal is to compute a set of top-$k$ words with the highest global impact according to different aggregation functions. Some of our techniques are lossless and some are lossy.
We show that for a very mild loss of quality, we are able to accelerate the computation by up to 30$\times$, reducing the computation from hours to minutes. We also devise and study a probabilistic model that accounts for noise in the Anchor algorithm and diminishes the bias toward words that are frequent yet low in impact.
\end{abstract}

\section{Introduction}

A particular paradigm for local explanations consists of algorithms that compute a numeric estimate of each token's contribution to the decision of the model, also known as input attribution~\cite{danilevsky-etal-2020-survey}. 
Many of the common techniques are computationally intensive. For instance, the score that Anchor~\cite{anchors:aaai18} assigns to a token is the probability that the model keeps its decision intact when the document undergoes random perturbations, as long as the token remains unchanged. LIME~\cite{DBLP:journals/corr/RibeiroSG16} derives a linear bag-of-words predictor of the model's outcome in a small area around the document and scores each token by its learned coefficient. The SHAP score~\cite{SHAP} views the tokens as players in the cooperative game of forming the model's decision, and applies the well-known Shapley value to attribute a portion of the profit to each player. As a running illustration, \Cref{fig:intro} shows documents from a spam-detection task of Amazon reviews on toys and games, with words marked as \e{anchors}, meaning that their score by Anchor exceeds a threshold.

In turn, an approach to global explanation is the \e{aggregation} of the attribution scores computed by methods as the aforementioned, intended to quantify the impact of each term on the model~\cite{DBLP:conf/rep4nlp/ArrasHMMS16}. 
More precisely, we begin with a set of instances to classify (e.g., the training set or the test set of a learning setup), compute a score for each token, and then estimate the impact of every word by aggregating the scores that it obtained in its occurrence in documents. \Cref{fig:intro} shows top-10 impactful words based on the anchors; to illustrate an insight, ``Christmas'' is one of the most impactful terms that encourage the model  \cite[Bert,][]{devlin2018bert} to classify a document as spam. Like any form of a global explanation, we would like it to be used effectively by users of varying skills to gain valuable insights on the model: what it learned in training, which biases it has towards specific phrases, which adversarial examples can shed light on possible weaknesses, and so on. For instantiation, we would like to present the top-$k$ impactful terms in online analysis frameworks such as the FIND system~\cite{DBLP:conf/emnlp/Lertvittayakumjorn20} that provides attribution-based insights.

\newcommand*{\bx}[1]{\underline{\textbf{#1}}}
\def\ttr#1{$\circ$\,\textbf{#1}\,\,}

\newenvironment{fancydoc}
{\begin{tcolorbox}[boxrule=1pt,width=1.15in, colframe=NavyBlue, boxsep=0mm, left=4pt,right=4pt,top=5pt,bottom=5pt]\flushleft\it
}
{\end{tcolorbox}}
\definecolor{lightyellow}{HTML}{FFFFE5}
\newenvironment{fancyterms}
{\begin{tcolorbox}[colback=white, arc=2pt,boxrule=1pt, width=1.7in, boxsep=0mm, left=4pt,right=4pt,top=4pt,bottom=4pt]\flushleft
}
{\end{tcolorbox}}

\begin{figure}
\parbox[t]{1.15in}{\centering\small
\vspace{-0.35cm}
\begin{fancydoc}
My daughter got this for \bx{christmas} and \bx{loves} it!
\end{fancydoc}
\vspace{-0.2cm}
{\large{$\boldsymbol{\vdots}$}}
\begin{fancydoc}
\bx{Great} service and shipping. Thank you. Because of you I had less stress for \bx{christmas}
\end{fancydoc}
Document collection: spam detection over Amazon toy reviews
}
\parbox[t]{0.25in}{
\vspace{0.2em}
\begin{tikzpicture}[ultra thick, baseline={([yshift={-\ht\strutbox}]current bounding box.north)},outer sep=0pt,inner sep=0pt]
\draw [decorate,  
    decoration = {brace,raise=0pt,aspect=0.75,amplitude=15pt,mirror}] (0,0) --  (0,3.0);
\end{tikzpicture}
}
\parbox[t]{1.8in}{\centering\small
\vspace{-1em}
\begin{fancyterms}
\textit{Top-10 terms:} \,
\ttr{great}  \ttr{love}  \ttr{game}  \ttr{fun} \ttr{christmas} \ttr{friends} \ttr{grandson}  \ttr{family} \ttr{favorite} \ttr{enjoy}
\end{fancyterms}
\includegraphics[width=1.7in]{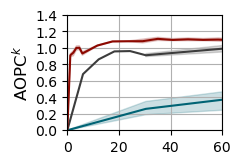}
\vskip-0.5em
Quality as a function of time (sec) for sum (\textcolor[HTML]{000000}{\textbf{---}}), 
avg (\textcolor[HTML]{0F6C7C}{\textbf{---}}), our (\textcolor[HTML]{8C0800}{\textbf{---}})
}
\caption{\label{fig:intro} Left: local explanations in a collection of documents for a spam detection task. Top-right: top-10 terms resulting from the global aggregation of (local) anchors. Bottom-right: global aggregation quality; our approach leads to higher quality at a fraction of the computation time. }
\end{figure}
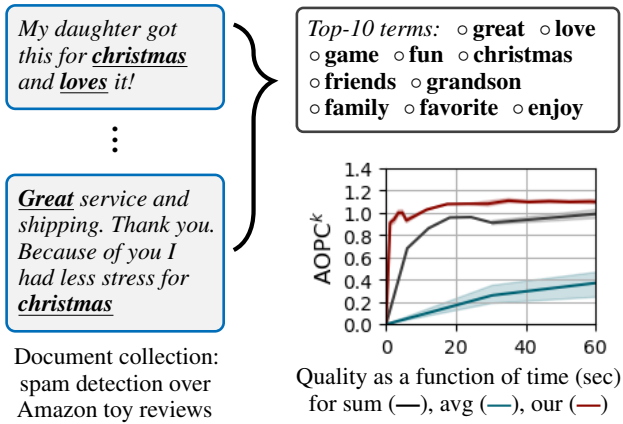

Yet, the global aggregation of a local attribution, like the above ones, entails critical challenges since the latter is designed to explain a single decision by the model. 
For one, how exactly should we aggregate the scores? The local scores are inherently noisy, and so, if we simply sum up the scores of all occurrences, we end up giving an advantage to frequent words that are irrelevant to the model (e.g., a function word like ``the'' is likely to receive enough noisy scores that sum up to a high global score). If we normalize by the number of occurrences, then we promote rare terms that matter only in insignificant cases (e.g., a word that only appeared once in the dataset would be scored highly). Either way, the aggregation falls short of the expectation to explain the model. 

Another major challenge, regardless of which aggregation is used, is the \e{computational cost}. The statistical approaches are arguably designed to be practical as local explanations, where we wish to provide an online answer for a single instance; yet, global aggregation should, in principle, apply the costly computation to every token of every document. For instance, a direct application of the release of Anchor on a collection of 10k documents took us almost two days, which were shortened to more than an hour after code optimizations. Again, such performance casts the usage of global aggregation as ineffective for online analysis pipelines. 

We focus on the global aggregation of the Anchor explanations and tackle the above challenges by making the following contributions. First, we design a probabilistic model that targets the importance of words \e{as explanations} and, at the same time, accounts for noise and irrelevant occurrences. Second, we develop several optimizations of global aggregation over Anchor towards an effective implementation. Some of the optimizations are specific to Anchor, while others apply to every global aggregation of contribution scores. Specifically, we design an \e{anytime} algorithm that maintains a set of top-$k$ candidates, and approaches the top score after a fraction of the na\"ive time, with an improvement of up to 30x. Third, we design and conduct an experimental study that shows the effectiveness of our solutions. For that, we adapt the evaluation method of  \citet{DBLP:journals/corr/abs-1802-01933} to the task of top-$k$ impactful terms. This is illustrated in \Cref{fig:intro}: after a few seconds, our anytime algorithm produced a set of words that outperforms mere sum and average (that terminate in ten minutes).

\paragraph*{Related work.}
The local attribution scores we discussed  fall in the category of \e{perturbation-based} explanations (Anchor) or \e{simplification-based} explanations (LIME, SHAP). An earlier perturbation-based explanation is occlusion~\cite{DBLP:conf/eccv/ZeilerF14}. 
Another common category is \e{gradient-based} explanations, such as Saliency \citep{Simonyan2014DeepIC}, InputXGradient \citep{Shrikumar2016NotJA}, and others  \cite{DBLP:conf/nips/PruthiLKS20,DBLP:conf/icml/SundararajanTY17}; these assume white-box access to the model, while perturbation-based methods can operate with any model in a black-box manner. 
See \citet{DBLP:conf/emnlp/AtanasovaSLA20} for a comparative analysis. 
Other scores are derived from an \e{attention mechanism} incorporated in the model~\cite{DBLP:conf/acl/LuZXLZX19,DBLP:conf/acl/XieMDH17}. Layer-wise Relevance Propagation (LRP) extracts from the network a linear model where, similarly to LIME, the coefficients are used as scores~\cite{DBLP:conf/rep4nlp/ArrasHMMS16}.

Aggregation of local attributions was studied by \citet{DBLP:journals/corr/abs-2211-05485}, who applied aggregation to the gradient-based local attribution of \citet{DBLP:conf/emnlp/BastingsEZSF22}.
Aggregation of LRP was done by 
\citet{DBLP:conf/emnlp/Lertvittayakumjorn20} and 
\citet{DBLP:journals/corr/abs-2106-07410}. 
These publications focused on representing the whole space of local attributions (e.g., via clustering, word clouds, and representative embeddings),  and did not focus on the challenge of execution cost. Later in the paper, we refer to work on the aggregation of LIME scores~\cite{DBLP:journals/corr/RibeiroSG16,DBLP:journals/corr/abs-1907-03039} that also did not focus on the computational aspects.

Token scoring is one of a plethora of explanation forms proposed for machine-learning models. Other popular approaches derive from the original model different kinds of insights such as nearby \e{interpretable} (or \e{surrogate})  models,  deterministic 
\e{rules}, and \e{examples} that highlight nuances of the model. These can be found in relevant surveys such as \citet{DBLP:journals/csur/GuidottiMRTGP19}, \citet{DBLP:conf/ijcnlp/DanilevskyQAKKS20} and \citet{DBLP:conf/emnlp/AtanasovaSLA20}, to name a few.

\def\docs{\mathbf{D}}
\def\dist{\Delta}
\def\prob{\mathrm{Pr}}
\def\anchors{\mathsf{Anc}}
\def\set#1{\mathord\{#1\}}
\def\G{\mathcal{G}}
\newcommand{\defeq}{\overset{\text{\tiny def}}{=}}
\def\gsqrt{\mathord{\G_{\mathsf{sq}}}}
\def\gave{\mathord{\G_{\mathsf{av}}}}
\def\gh{\mathord{\G_{\mathsf{h}}}}
\def\gpr{\mathord{\G_{\mathsf{pr}}}}
\def\tgpr{\mathord{\tilde{\G}_{\mathsf{pr}}}}
\def\aopcg{\mathsf{AOPC}_{\mathsf{global}}}
\def\tG{\tilde{\G}}

\section{Formal Setup}\label{sec:setup}
A \emph{document} $d$ is a finite sequence $w_1,\dots,w_m$ of \emph{words}. For $i=1,\dots,m$, we call the pair $(w_i,i)$ 
a \emph{token} of $d$. By a slight abuse of notation, we may identify $d$ with its set of tokens; hence, $(w,i)\in d$ means that the $i$'th word of $d$ is $w$. 
We denote by $\docs$ the set of all documents. By a  \emph{predictor} we refer to any function $f:\docs\rightarrow C$ that maps documents to some domain $C$. In particular, a binary classifier is a predictor where $C=\{0,1\}$. In our evaluation, we will make the (conventional) assumption that $f(d)$ is the most likely class according to an associated probability function $\hat f:\docs\times C\rightarrow[0,1]$  that defines a distribution over $C$ for each document $d$.

\paragraph{Anchors.}
The anchor concept has been defined for general modalities and ``predicates'' (properties of the input instance) that can take the role of an explanation of a prediction~\cite{anchors:aaai18}. We define it in the textual domain where the predicates are token memberships. 

We assume that every document $d$ is associated with a \emph{perturbator}, which is a distribution $\dist_d$ over $\docs$.
We later discuss the actual perturbation used in the public anchor implementation. 
We also assume a numerical threshold $\tau\in[0,1]$. Let $f$ be a predictor, and let $(w,i)\in d$ be a token of $d$. We say that $(w,i)$ is an \emph{anchor} (of $d$ w.r.t.~$f$) if
\begin{equation}
 \label{eq:anchor}
 \prob_{x\sim \Delta_d}[f(x)=f(d) \mid (w,i) \in x]\geq\tau\,.
\end{equation}
 In words, $(w,i)$ is an anchor if the document retains the same prediction under perturbation, with high probability, as long as the word $w$ is kept at the $i$th position in the perturbation. This probability is referred to as the \emph{precision} of $(w,i)$. We denote by
$\anchors(d)$ the set of anchors of a document $d$.

The algorithm uses a Masked Language Model (MLM) as the perturbator of $d$. Samples from $\docs$ are created by masking tokens of $d$, 
and later unmasking using DistilBert~\cite{DistilBert}.
The output of the MLM on a masked $(w,i)$ is a distribution of words that can fit the mask. Out of the $\zeta$ (500 here) words with the highest probability, a word is sampled according to its probability. If there are multiple masks, then the process repeats iteratively until all replacements are applied.

\subsubsection{Global aggregation.}
\label{sub_sec:global}
Let $S$ be a finite collection of documents, and let $f:\docs\rightarrow C$ be a predictor that we wish to explore in the context of $S$. (For instance, $S$ can be the training set that is used for the construction of $f$.) We denote by $W(S)$ the set of words that occur in $S$. 
By \emph{global aggregation} we refer to any numerical function $\G$ that maps every word $w\in W(S)$ and $c\in C$ to a number $\G(w,c)$. Intuitively, high 
$\G(w,c)$ means that $w$ has high impact on $f$ taking the value of $c$ on a given document.
We denote by $S[c]$ the collection of documents in $S$ classified as $c$. We also denote by $A^+(w,c)$ and $A^-(w,c)$ the number of occurrences of a word $w\in W(S)$ where $w$ is considered an anchor for $c$ and a non-anchor for $c$, respectively:
\begin{align}
 &A^+(w,c) \defeq\sum_{d\in S[c]} |\{i \mid (w,i) \in \anchors(d) \}|\label{eq:aplus}\\
 &A^-(w,c) \defeq \sum_{d\in S[c]} |\{i \mid (w,i)\in d\setminus \anchors(d)\label{eq:aminus}\}| 
\end{align}
\citet{DBLP:journals/corr/RibeiroSG16} proposed the function:
\begin{align}\label{eq:gsqrt}
    \gsqrt(w,c) \defeq 
    \sqrt{A^+(w,c)}
\end{align}

Several aggregations have been proposed by \citet{DBLP:journals/corr/abs-1907-03039}. 
The one that performed best on binary classification is the \emph{Global Average Importance} that, in our context, is normalizing $A^+(w,c)$ by the number occurrences of $w$:
\begin{equation}
\label{eq:gave}
\gave(w,c)\defeq 
\frac{A^+(w,c)}{A^+(w,c) + A^-(w,c)}
\end{equation}
Note that $\gave(w,c)$ does not distinguish between rare words that occur as anchors and words that are frequently anchors. In fact, the maximal score is obtained by a word that occurs once, and in that occurrence, it is an anchor. On the other hand, $\gsqrt(w,c)$ is sensitive to the noise of the anchor algorithm since it rewards frequent words (e.g., stop words) that are occasionally identified as anchors. 

Another aggregation that \citet{DBLP:journals/corr/abs-1907-03039} proposed is $\gh(w,c)$, which weighs a word by its score for different classes. The idea is that the multiplicity of classes entails a penalty since it indicates low relevance to the specific class.
Let $\widetilde{h}(w,c)$ be the normalized $\gsqrt(w,c)$, that is, 
$\widetilde{h}(w,c) \defeq \gsqrt(w,c)/\sum_{c' \in C} \gsqrt(w,c')$.
Note that $\widetilde{h}(w,c)$ can be viewed as a distribution of $w$'s importance across the classes. The Shannon entropy of this distribution is 
$H_w \defeq  - \sum_{c \in C} \widetilde{h}(w,c) \log \left(\widetilde{h}(w,c) \right)$.
Low $H_w$ implies that $w$ impacts a specific class. Let $H_{\min}$ and $H_{\max}$ be the minimum and maximum $H_{w'}$ across all words $w'$.
Since rare words might occur in a specific class (thus having low entropy), $\gh$ is defined using $H_w$ and $\gsqrt$ as follows:
\begin{equation}\label{eq:gh}
    \gh(w,c) \defeq  \left( 1 - \frac{H_w - H_{\min}}{H_{\max} - H_{\min}} \right) \gsqrt(w,c)
\end{equation}
Hence, a high entropy will have a low factor over $\gsqrt$, and rare words will have low $\gsqrt$, both resulting in low $\gh$.

While $\gh(w,c)$ aims to address the problems of $\gave(w,c)$ and $\gsqrt$, we have found that anchors commonly appear in at most one class, thus the entropy is very low for most words to begin with. In \Cref{sec:probmodel}, we will propose a new aggregation that aims to overcome these weaknesses.

\subsubsection{Problem definition: top-k terms.}\label{Problem def}
We consider the following computational task. We have a set $S$ of documents, a predictor $f$, an aggregation function $\G$ for anchors, and a number $k$. The goal is to find, for a given class $c$ of $f$, a set $T_c$ of $k$ words in $W(S)$ with maximal $\G(w,c)$.  We will refer to these words as the \emph{top terms}.
While we wish to find $T_c$ in interactive time, a naive aggregation can be prohibitively slow. We might be willing to settle for an approximation, that is, a set $T'_c$ of $k$ words has \emph{similar quality} compared to $T_c$. Next, we discuss how this quality can be measured.

For evaluating global aggregations, \citet{DBLP:journals/corr/abs-1907-03039} proposed $\aopcg$ that adapts the \emph{Area Over the Perturbation Curve} of~\citet{DBLP:journals/corr/SamekBMBM15}. The idea is to measure how removal of high-score terms impacts the model's predictions. In our terms, $\aopcg$ is calculated on an aggregation $\G$ by progressively removing from each document the top-$k$ terms $w$ in that document by decreasing $\G(w,c)$. A curve shows a better performance than another if it is higher and its initial slope is steeper (indicating better identification of the top influencing terms).

Our goal is to measure the quality of the set $T_c$ of top-$k$ terms. Hence, in our experiments, we adapt $\aopcg$ so that we remove from each document $d$ \emph{only the terms in $T_c$}; in particular, documents that do not intersect with $T_c$ remain unchanged (in contrast to $\aopcg$ that removes $k$ terms from each document). 
Formally, let $f:\docs\rightarrow C$ be a predictor, $c\in C$ a class, and $T_c$ the set of top-$k$ terms found by the aggregation $\G$ and a class $c$. Let $w_1,\dots,w_k$ be the words in $T_c$ ordered by decreasing $\G(\cdot,c)$.
For a document $d$, let $d^i$ be the document $d$ with every occurrence of a word from $w_1,\dots,w_k$ removed.
We define
\begin{equation}
\aopck(\G,c) \defeq \frac{1}{k + 1}\cdot\underset{d\in S[c]}{\mathrm{avg}}\Big(
\sum_{i=1}^k \hat f(d,c) - \hat f(d^i,c)\Big)
\label{eq:aopck}
\end{equation}
where $\hat f$ is the probability that the classification model of $f$ assigns to $c$ for the document $d$. 
In words, $\aopck(\G,c)$ is the average drop in the probability of the class $c$, over all documents and prefixes of $T_c$.

\section{Probability-Based Global Aggregation}
\label{sec:probmodel}
Consider a document collection $S$, a predictor $f:\docs\rightarrow C$, and a class $c\in C$. Let $d=(w_1,\dots,w_\ell)$ be a document in $S[c]$. We consider a simple generative model that produces a random document $X_d=(w'_1,\dots,w'_\ell)$ of the same length as $d$, assuming that each word is generated randomly, yet anchor words and non-anchor words are taken from different distributions. More precisely, each $w'_i$ is selected randomly and independently using the following process: 
\begin{itemize}
\item If $(w_i, i) \in\anchors(d)$, then do as follows. With probability $\alpha$, select a word $w\in W(S)$ with the \emph{anchor probability} $q(w,c)$;
with probability $1-\alpha$ (that the anchor is wrong), select a word $w$ from $W(S)$ with the probability $p(w, c)$.
\item Otherwise, select $w\in W(S)$ with probability $p(w, c)$.
\end{itemize}
Here, $\alpha$ is a parameter. The probabilities $p(w, c)$ and $q(w,c)$ are unknown and chosen to maximize the probability of $S$:
\begin{equation}\label{prob formula}
    (p^*,q^*) \defeq \argmax_{p,q}\Big(\prod_{d\in S[c]}\prob[d=X_d]\Big)
\end{equation}

Intuitively, words $w$ with high $q^*(w,c)$ are likely to be used as anchors but not necessarily as non-anchors. Hence, we use $q^*$ as our global aggregation: 
\begin{equation}
    \label{eq:pr}
    \gpr(w,c)\defeq q^*(w,c)
\end{equation}

We estimate $q^*$ and $p^*$ via Maximum Likelihood Estimation using Lagrange multipliers~\cite{SARGENT2000361} for local maxima. Our analysis is in the Appendix for space limitations. The resulting estimations $\tilde{q}$ and $\tilde{p}$ are as follows. Recall that $\alpha$ is a parameter, and recall $A^+(w,c)$ and $A^-(w',c)$ from Equations~\eqref{eq:aplus} and~\eqref{eq:aminus}, respectively. 
\begin{align}
    &\tilde{q}(w, c) = \frac{1}{\alpha}\cdot\dfrac{A^+(w,c)}{\Sigma_{w'}A^+(w',c)} - (\frac{1}{\alpha}-1)\cdot\dfrac{A^-(w,c)}{\Sigma_{w'}A^-(w',c)}\notag\\
    &\tilde{p}(w, c) = \dfrac{A^-(w,c)}{\Sigma_{w'}A^-(w',c)}\label{eq:pqtilde}
\end{align}

The resulting $\tilde{q}$ is not necessarily the equal $q^*$ since it is not necessarily positive. To get a probability (which may be different from $q^*$), we apply \emph{Laplace smoothing}~\cite{CHEN1999359} with the absolute minimum value of $\tilde{q}$ as the smoothing parameter. Let
$\qmin \defeq \min_{w\in W(S[c])}{\tilde{q}(w, c)}$
and $\mathcal{L}(g, \beta)$ be the Laplace smoothing over the function $g$ with a smoothing parameter $\beta$.
We then define $
    \tilde{q}^*(w,c) = \mathcal{L}(\tilde{q}, |\qmin|)$ and 
    $\tilde{p}^*(w, c) = \tilde{p}(w,c)$.
Laplace smoothing is monotone, thus it preserves word ordering: $\tilde{q}^*(w_1, c) \geq \tilde{q}^*(w_2, c)$ whenever $\tilde{q}(w_1, c) \geq \tilde{q}(w_2, c)$.

\section{Runtime Optimizations}\label{sec:optimizations}
In this section, we propose algorithms and optimization techniques for accelerating the computation of the top-$k$ terms.

\subsection{Incremental Evaluation (Anytime)}\label{subsec:incremental}
Our goal is to retrieve the top-$k$ words in the dataset. Thus, we adopt an incremental evaluation that maintains the best top-$k$ found in each step of the algorithm. This allows us to provide the user with informative early results in a \emph{pay-as-you-go} (\emph{anytime}) manner. Yet, some of the scores are not known during the computation, since they may require the processing of the entire dataset. Hence, we use a heuristic \emph{pseudo-score} $\tG(w, c)$ instead of each exact aggregation $\G(w,c)$.

We traverse the documents in a predefined order (that we discuss next). Denote the word with the lowest pseudo-score in the top-$k$ group as $\wmin$.
A word $w$ enters the top-$k$ group if $\tG(w, c) > \tG(\wmin, c)$. 
Since the results of the anchor algorithm are unknown for all documents during its running, the pseudo-score when reaching document $d_i$ is calculated with respect to only the first $i$ documents. Specifically, denote by $A^+_i(w,c)$ and $A^-_i(w,c)$
the values $A^+(w,c)$ and $A^-(w,c)$, respectively, restricted to the first $i$ documents in $S[c]$.
For $\gsqrt, \gave$ and $\gh$, the pseudo-score is calculated simply by replacing $A$ with $A_i$ in the definitions. In the case of $\gpr$, we use the estimates $\tilde{q}$ and $\tilde{p}$ of \Cref{eq:pqtilde} and, again, apply the replacement of $A$ with $A_i$. 
As for the order, we traversed the documents by descending confidence of the classification, assuming that these encourage anchors early on. (We also tried other orders and got similar results.)

\subsubsection{Candidate filtering.}
By utilizing the maintained set of top candidates, we can avoid the expensive computation for an unlikely candidate, by applying an optimistic, fast-to-compute upper bound on its score: a word is filtered out if its upper bound is lower than the minimal pseudo-score in the current set of candidates. The upper bound for a word $w$ considers the occurrences $(w,i)$ of the word in the remaining documents $d$, and it is the pseudo-score under the assumption that $(w,i)\in\anchors(d)$ for every such $d$ and $i$.
While this bound might appear optimistic, this optimization accounts for 30\%-40\% reduction of the computation's running time.

\subsection{Accelerating Anchor}\label{sec:accelerate}
Recall that a major challenge in computing the top terms is the expensive computation of anchors, as discussed in \Cref{Problem def}. Yet, we are interested in an aggregation over many applications of the Anchor algorithm, and so, we might settle for lower accuracy and confidence. Hence, we change the hyperparameters of the Anchor algorithm, as follows. 

\paragraph{Masking.}
Recall that Anchor generates documents by masking words in the instance and filling them using a DistilBert MLM model. Each mask is filled out of $500$ MLM suggestions. We observed that \emph{decreasing} this number (to just 50 in our experiments) reduced the execution cost without harming the quality, and sometimes even improving it.

\paragraph{Anchor precision and confidence.}
The precision threshold $\tau$ of \Cref{eq:anchor} determines the number of documents that we sample using the replacement alternatives produced by the masking. During runtime, we reduce $\tau$ for a word to reflect its past occurrences as an anchor. Consequently, we generate fewer documents to decide whether the word is an anchor. Instead of $\tau$, we use 
\begin{align}\label{eq:confidence}
    \tilde{\tau}(w, c) = \tau - \omega \cdot \tgpr(w, c)/{N_w}
\end{align}
where $N_w$ is the number of occurrences of $w$ in the dataset, and 
$\omega$ is a hyperparameter. We use $\omega=0.4$ in our experiments so that
 $\tilde{\tau}(w, c)$ does cannot get below $0.55$.

In addition, the anchor implementation by \citet{anchors:aaai18}
estimates, for each anchor candidate $w$, a \emph{confidence score} that the inequality of
\Cref{eq:anchor} is correct.
To be an anchor, this confidence needs to exceed $1-\delta$. Higher values of $\delta$ lead to a reduction in the number of needed samples. Hence, since we aggregate results, we can settle for higher values of $\delta$.

\begin{figure*}[t]
    \includegraphics[width=1.0\linewidth]{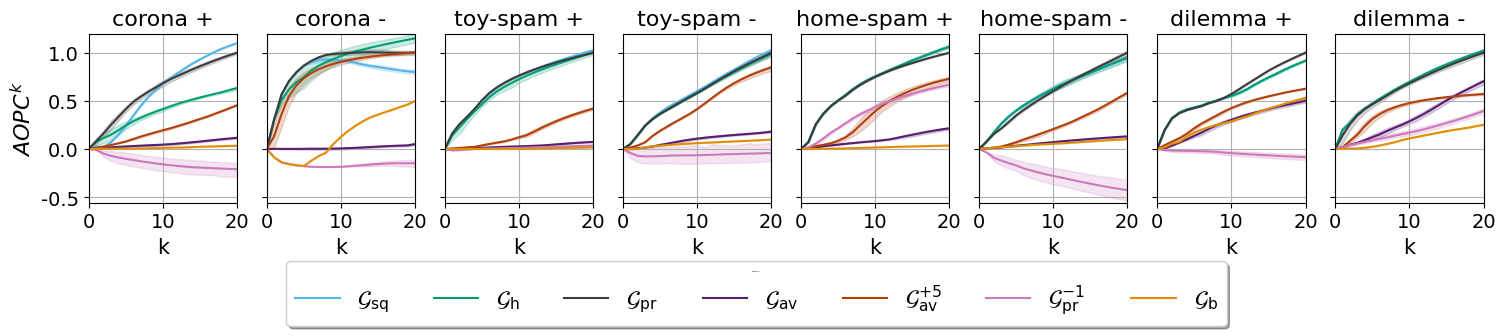}
    \caption{$\aopck(\G,c)$  of different aggregation functions $\G$ (y-axis) for varying $k$ (x-axis).}
    \label{fig:Deberta aggregation}
\end{figure*}

\begin{figure*}[t]
    \includegraphics[width=1.0\linewidth]{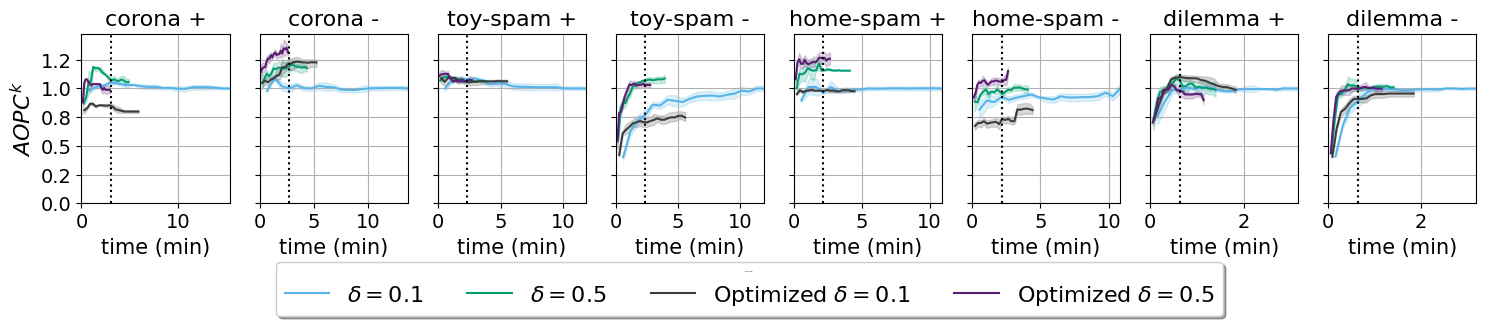}
    \caption{$\aopck(\G,c)$ for different versions $\G$ of $\gpr$ and $k=20$, as a function of the computation time. ``Optimized'' refers to the combination of all optimizations described in \Cref{sec:optimizations}.}
    \label{fig:optimization-bottomline}
\end{figure*}

\subsection{Additional Optimizations}\label{subsec:additional}
In addition to the optimizations described in this section, we applied several accelerations that are fairly straightforward and standard. 
First, we filtered out \emph{stop-words} from the set of top-term candidates, as we perceive them as non-informative terms regarding the explanation of the model. We also filtered out \emph{rare words}, where we defined a word to be rare if it occurs fewer times than some threshold (5 in our experiments). Our experiments confirm that this filtering provides considerable acceleration for negligible loss of quality. Finally, we also experiment with the application of the entire algorithm on a \emph{sample} of the documents rather than the entire training set.

\section{Experiments and Results}
In our experimental study, we aim to understand the effectiveness of the aggregations and optimizations that we proposed. More precisely, we investigate, empirically, how well and how fast each alternative finds top-$k$ terms. The code and data are available at \url{https://github.com/alonm16/anchor}.

\subsection{Setup}
We find the top-$k$ terms for several classification tasks. We use $k=20$. In each task, 
the documents are organized into three collections: training, validation, and test. The predictor $f$ is a classifier trained on the training set, chosen as the model checkpoint with the best validation set accuracy. We experimented with three models: logistic regression, Bert~\cite{devlin2018bert}, and DeBERTa~\cite{deberta}. We report the results only for DeBERTa; the other two models are consistent with those of this section and are given in the Appendix. The set $S$ of documents that we aggregate over (as described in \Cref{sec:setup}) is the test set, as experimented in \citet{DBLP:journals/corr/abs-1907-03039}.

All experiments were conducted on a machine with $96$ of Intel Xeon Gold 6336Y 2.40GHz CPUs with $24$ cores, 512GB RAM, $8$ of 50GB Nvidia A40 GPUs running Ubuntu 20.04 LTS. The algorithms were programmed in Python 3.10 with the libraries CUDA 11.6, PyTorch 2.0, and Numpy 1.23.

\subsubsection{Classification tasks.}

We restricted each dataset to documents of at most $200$ characters, since the influence of each token on a longer text diminishes, and we got fewer anchors and less meaningful results.\footnote{The length bound also influences the execution cost; our experiments show that the runtime grows linearly with this bound.}
We used the following tasks.

\def\dset#1{\smallskip
\noindent
\underline{#1}}

\dset{Coronavirus tweets (sentiment)}.\footnote{\scriptsize\url{https://www.kaggle.com/datasets/datatattle/covid-19-nlp-text-classification}}
Tweets classified into five sentiments: extremely negative, negative, neutral, positive, and extremely positive. We combined extremely negative and negative, and extremely positive and positive since there were too few anchors that distinguish between the extreme and normal classes. The dataset consists of 16,000, 4200, and 12,500 documents (training, validation, test).

\dset{The Social Dilemma tweets (sentiment)}.\footnote{\scriptsize\url{https://www.kaggle.com/datasets/kaushiksuresh147/the-social-dilemma-tweets}}
Tweets classified into three sentiments: positive, negative, and neutral. The dataset consists of $3200$, $1000$, and $3000$ documents.

\begin{figure*}
    \includegraphics[width=1.0\linewidth]{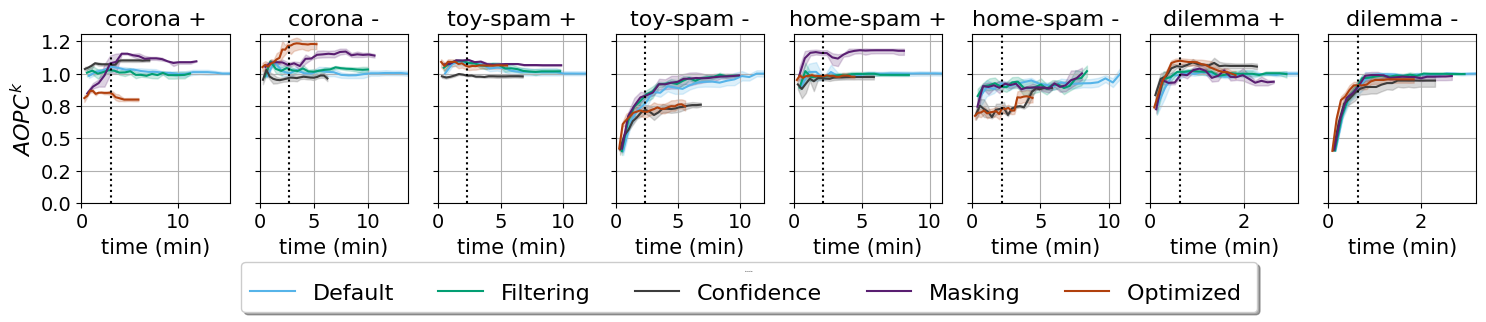}
    \caption{$\aopck(\G,c)$ for different optimizations of $\gpr$ and $k=20$, as a function of the computation time. ``Optimized'' refers to the combination of all optimizations.\label{fig:optimization-sep}}
\end{figure*}

\begin{figure*}[ht]
    \centering
    \includegraphics[width=1.0\linewidth]{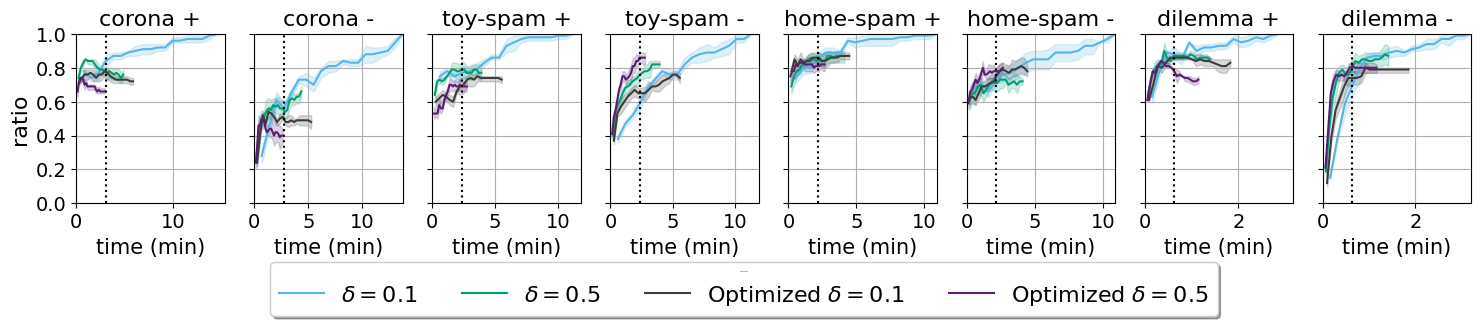}
    \caption{Ratio of shared terms for different versions of $\gpr$.\label{fig:time percent optimizations}}
\end{figure*}

\dset{Amazon reviews: Toys \& Games (spam)}.\footnote{\scriptsize\url{https://www.kaggle.com/datasets/naveedhn/amazon-product-review-spam-and-non-spam?select=Toys_and_Games}}
Amazon product reviews in the Toy and Games category. The reviews are classified into \emph{spam} and \emph{non-spam}. The dataset consists of 15,000, 3800, and 11,000 documents.

\dset{Amazon reviews: Home \& Kitchen (spam).}\footnote{\scriptsize\url{https://www.kaggle.com/datasets/naveedhn/amazon-product-review-spam-and-non-spam?select=Home_and_Kitchen}}
Similar to the previous one, but now in the category Home and Kitchen. The dataset consists of 12,000, 3000, and 9000 documents.

\subsubsection{Compared Aggregations.}\label{sec:compared-aggs}
We compare the aggregation functions of Sections~\ref{sec:setup} and~\ref{sec:probmodel}, and the optimized versions of \Cref{sec:optimizations}. The list of functions includes $\gsqrt$ (\Cref{eq:gsqrt}), $\gave$ (\Cref{eq:gave}), $\gh$ (\Cref{eq:gh}), and $\gpr$ (\Cref{eq:pr}) with $\alpha=0.5$. As for $\gave$, we also experiment with a \emph{quick fix} of its weakness of promoting rare words: the function $\gavemod$ is the same as $\gave$, except that every word is filtered out if it has fewer than five occurrences in the training set. 

As baselines, we use two aggregations $\G(w, c)$. The first ignores the Anchor algorithm and simply measures the probability of class $c$ among the documents that include $w$.
\begin{equation}\label{eq:gb}
    \gb(w, c) \defeq \dfrac{|\set{d\in S[c]\mid w\in d}|}{|\set{d\in S\mid w\in d}|}
\end{equation}
(Formally, $w\in d$ means $(w,i)\in d$ for some $i$.) The second is $\grev\defeq 1/\gpr$, the inverse of $\gpr$, used as a sanity check.

\subsection{Results}\label{subseq:results}
\Cref{fig:Deberta aggregation} shows the results for the aggregation functions. 
    Each box corresponds to one task and one class $c$, and it shows $\aopck(\G,c)$ for the top terms at different times. 
Each line includes a shaded error band based on repetitions with 5 seeds. Observe that $\gpr$ consistently begins with the steepest curve, and its overall height exceeds the baselines. 
Also note that $\gave$ is inferior to the rest. In contrast, $\gsqrt$ and $\gh$ perform similarly to $\gpr$. While so, in \Cref{sec:case-studies} we inspect the three in several case studies, and argue that the results of $\gpr$ are more useful, even though it is not captured by $\aopck$.

Figures~\ref{fig:optimization-bottomline} and~\ref{fig:optimization-sep} show the quality of $T_c$ when applying the aggregation in an anytime manner. We measure the quality of the set of $k$ candidates at different times during the computation, until completion. Hence, each chart shows the change of $\aopck$ over time as well as the total running time. 

\Cref{fig:optimization-bottomline} shows that increasing $\delta$ improves \emph{both the running time and quality} of $\gpr$. 
An exception is ``dilemma +'' where we get reduced runtime but a slight decrease in quality.
This suggests that the aggregation compensates for the reduction in confidence (number of samples) for the Anchor algorithm. It also confirms the conjecture that the default value of $\delta$, namely $0.1$, is too strict within aggregation. Combining the remaining optimizations shortens the overall running time; 
while it reduced the quality in the ``corona +'' experiment, it generally did not impair and sometimes improved the quality (e.g., ``corona -'' and ``home-spam -/+''). This experiment shows that, overall, the optimizations are highly beneficial.

\Cref{fig:optimization-sep} shows that each optimization alone accelerates the computation. The \emph{confidence} optimization (\Cref{eq:confidence} in \Cref{sec:accelerate}) incurs the most significant reduction in quality. In contrast, the \emph{masking} optimization (\Cref{sec:accelerate}) consistently improves the overall quality. Applying all optimizations together yields the shortest running time, but generally at the cost of lowering the $\aopck$ scores (with the exception of ``corona -'' where the impact on the score is positive). 

In
\Cref{fig:time percent optimizations}, we measure the percentage of shared terms between $T_c$ of $\gpr$ to that of its optimizations. The intersection is generally around $80\%$. Note that a drop in the intersection does not necessarily imply a drop in quality, since different terms can have similar quality (as shown in Figures~\ref{fig:optimization-bottomline} and~\ref{fig:optimization-sep}). 

\subsection{Case Studies}
\label{sec:case-studies}
We now discuss several case studies from our experiments. 
\Cref{tab:toy-spam} shows the top terms for the toy-spam dataset under $\gpr$. We can see that terms for spam reviews are mostly positive (promoting products). On the other hand, non-spam reviews use more negative adjectives and commonly include customer complaints. (As an exception, ``thought'' occurs in non-spams in $80\%$ of the cases, and is typically used negatively to describe disappointment from a product.)

\begin{table}[t]\small
    \begin{tabular}{p{0.2\textwidth} p{0.2\textwidth}}
        \multicolumn{1}{c}{toy-spam -} & \multicolumn{1}{c}{toy-spam +}\\
    \midrule
        small, broke, disappointed, waste, would, smaller, cheap, thought, money, poor & great, love, loves, fun, favorite, awesome, wonderful, loved, classic, perfect\\ 
    \bottomrule
    \end{tabular}
    \caption{Top-10 terms: negative vs.~positive (no stop words).}
    \label{tab:toy-spam}
\end{table}

\Cref{table:casestudy-top} shows the results for the toy-spam (negative) and home-spam (positive) datasets. We can see that $\gsqrt$ has many common insignificant words, such as ``my,'' ``it,'' ``and,'' ``the,'' and so on. The function $\gave$ selects rare words such as ``mouths,'' ``files,'' ``visited,'' and ``wars.'' The function $\gpr$ is balanced and selects different terms such as ``wonderful'' and ``happy'' (as positives) or ``broke'' and ``waste'' (as negatives).
\begin{table}\small
    \begin{tabular}[t]{p{0.20\textwidth}}
    \multicolumn{1}{c}{toy-spam -}\\
    \toprule
        $\gpr$:
        not, but, disappointment, broke, small, it, would, waste, only, price \\
        \midrule
        $\gave$:
        narrower, tipping, expanding, mouths, stains, tone, files, obscure, faint, health\\ 
        \midrule 
        $\gsqrt$:
        not, but, it, the, disappointment, to, this, broke, was, small\\
    \bottomrule
    \end{tabular}
\,
    \begin{tabular}[t]{p{0.20\textwidth}}
    \multicolumn{1}{c}{home-spam +}\\
    \toprule
        $\gpr$:
        great, love, best, excellent, perfect, loves, wonderful, happy, pleased, good  \\ \midrule
    $\gave$:
    castle, boys, visited, wars, kumar, vibrant, lavender, implement, farewell, silky
    \\ \midrule
    $\gsqrt$:
    great, love, best, good, and, my, I, perfect, loves, ever\\
    \bottomrule
    \end{tabular}
    \caption{Comparing different aggregations of the toy-spam and home-spam datasets.}
    \label{table:casestudy-top}
\end{table}

\Cref{tab: corona indices} shows the top terms of the corona dataset for $\gpr$ and $\gsqrt$, each with its position for the other aggregation. We can see, for example, that indices of common words like ``corona''  and ``19'' are dropped for $\gpr$, while less common words that appeared more as anchors stay at the top.  As insights on the dataset, the reader can see that ``hand'' and ``help'' are significant to the positive class, where the former is typically used in the context of hand sanitizing.

\begin{table}\small
    \begin{tabular}{p{0.44\textwidth}}
        \toprule
        $\gpr$:
        hand, like, help, good, safe, please, thank, great (0-7),
        free (9),
        support (11),
        thanks (12),
        well (13),
        best (15),
        positive (18),
        better (21),
        care (20),
        love (22),
        safety (25),
        relief (26)\\
        \midrule
        $\gsqrt$:
        hand, like, help, good, please, safe, thank, great (0-7), {19} (48), free (8), {co} (751), support (9), thanks (10), well (11), {corona} (752), best (12), {store} (33), {grocery} (42), positive (13)\\
        \bottomrule
    \end{tabular}
    \caption{Top terms of the corona dataset under $\gpr$ and $\gsqrt$. Each term is attached the position by the other function.}
    \label{tab: corona indices}
\end{table}

\paragraph{Counterfactual examples.}
In \Cref{tab:accuracy-drop-neutral}, we ran the following experiment, inspired by the work of \citet{wallace-etal-2019-universal} on the impact of concatenated text on the model's performance. For various tasks and classes, we manually generated short sentences with their top-10 terms of $\gpr$ (Optimized). Each sentence is appended to all documents with an opposite label. We then measured the drop in the overall accuracy of the classifier due to the change; that is, we compared the accuracy before to after the change. We then repeated the measurement when replacing the term with another word of a similar nature. Importantly, the word replacement is such that the meaning of the sentence should barely impact that classification, so one could expect similar drops.  Nevertheless, we can see that applying this change reduces the accuracy by a considerable amount ($\sim50\%$). This suggests that our method indeed finds significant terms for the model. 

\begin{table}[t]\small
    \centering
    \begin{tabular}{lp{0.53\columnwidth}l}
        \toprule
        Dataset & Addition & Drop (\%)\\  
        \midrule
        toy-spam - & I was a bit (\textbf{disappointed} / unsatisfied) with this game's performance. & 32/2 \\  \midrule
        toy-spam - & It is just a (\textbf{small} / usual) item. & 60/8 \\ \midrule
        toy-spam + & This store contained (\textbf{classic} / board) games. & 52/0 \\ \midrule
        toy-spam + & That game's theme is (\textbf{love} / animals). & 44/6\\\midrule
        corona + & I bought (\textbf{hand} / -) sanitizers. & 44/2\\\midrule
        corona + & People should (\textbf{support} / back) others more. & 62/1\\ \midrule
        corona - & People shouldn't fret over this (\textbf{crisis} / situation). & 62/2 \\\midrule
        corona (-) & The pandemic affected (\textbf{crude} / -) oil prices. & 57/7 \\
        \bottomrule
    \end{tabular}
    \caption{Sentences with top-10 anchors by $\gpr$ (in bold). Each sentence is appended to all documents with the opposite label. The average accuracy is shown before and after the change.}
    \label{tab:accuracy-drop-neutral}
\end{table}

\section{Conclusions}
We studied the problem of identifying the top-$k$ terms under an aggregation of their identification as anchors or non-anchors in the dataset. We proposed the probabilistic aggregation $\gpr$ as a way of accounting for both the frequency of words and their treatment likelihood of being anchors. Global aggregation over the Anchor explanations incurs a prohibitive computational cost. We proposed techniques for considerably accelerating the identification of the top-$k$ terms and showed experimentally that we obtain an anytime solution that is much more useful for online analysis, reducing the time from hours to minutes and seconds. 
We focused on single-word terms, and it is left for future work to study the case of multiple words, where the challenge is bigger due to the number of candidates for the top-$k$ terms. Finally, while the runtime optimizations were applied to Anchor, the general framework can be adapted to any method of local attribution scores, and such adaptations are also left for future work.

\section*{Acknowledgments}
This research was supported by the German-Israeli Foundation for Scientific Research and Development (grant I-1502-407.6/2019), 
the German Research Foundation (project 412400621),
the Israel Science Foundation (grant 448/20), an Azrieli Foundation Early Career Faculty Fellowship, and an AI Alignment grant from Open Philanthropy. 

\bibliography{main}

\appendix
\onecolumn

\section{Estimation of the probabilistic model}
\label{sec:appendix}
Our objective is $\argmax_{p,q}\left(\prod_{d\in S[c]}\prob[d=X_d]\right)$:
\begin{align*}
    &\argmax_{p,q}\left(\prod_{d\in S[c]}\prob[d=X_d]\right) = \argmax_{p,q}\left(\prod_{d\in S[c]} \prod_{w \in d} \prob[w=w'] \right) = 
    \argmax_{p,q}\left(\sum_{d\in S[c]} \sum_{w \in d} \log \prob[w=w'] \right)\\
    &=\argmax_{p,q}\left( \sum_{d\in S[c]} \biggl( \sum_{w \in \anchors(d)} \bigl(\log \prob[w=w']\bigl) + \sum_{w\in d \setminus \anchors(d)} \bigl(\log \prob[w=w']\bigl)\biggl)\right)\\
    &=\argmax_{p,q}\left( \sum_{d\in S[c]} \biggl( \sum_{w \in \anchors(d)} \Bigl(\log \bigl(\alpha\cdot q(w, c) + (1-\alpha)\cdot p(w, c) \bigl) \Bigl) + \sum_{w\in d \setminus \anchors(d)} \Bigl(\log p(w, c) \Bigl)\biggl)\right)\\
    &=\argmax_{p,q}\left(\sum_{w\in W(S[c])} A^+(w, c) \cdot \log (\alpha\cdot q(w, c) + (1-\alpha) \cdot p(w, c)) + A^-(w, c) \cdot \log p(w, c) \right) 
\end{align*}

\eat{
Our objective is $\underset{\theta}{\argmax} \, p(\vec y)$:
\begin{align*}
    \underset{\theta}{\argmax} \, p(\vec y) &= \underset{\theta}{\argmax} \prod_{i=1}^M p(d_i, a_i) = \underset{\theta}{\argmax} \prod_{i=1}^M p(d_i| a_i)\cdot p(a_i) = \underset{\theta}{\argmax} \prod_{i=1}^M p(a_i) \prod_{i=1}^M p(d_i| a_i) = \\
    &\underset{\theta}{\argmax} \prod_{i=1}^M p(d_i| a_i) \underset{(2)}{=} \underset{\theta}{\argmax} \prod_{i=1}^M \prod_{j=1}^N p(w_i^{j}| a_{i}) \underset{(3)}{=} \underset{\theta}{\argmax} \prod_{i=1}^M \prod_{j=1}^N p(w_i^{j}| a_i^{j})\\&= \underset{\theta}{\argmax} \sum_{i=1}^M \sum_{j=1}^N \log p(w_i^{j}| a_i^{j}) 
    = \underset{\theta}{\argmax} \bigl(\sum_{k=1}^K \#_{1w_k}\log p(w_{k}| 1) + \#_{0w_k}\log p(w_{k}| 0)\bigl) \\
    &= \underset{\theta}{\argmax} \bigl(\sum_{k=1}^K \#_{1w_k}\log (\alpha\cdot \theta_1 [k] + (1-\alpha)\cdot \theta_0 [k]) + \#_{0w_k}\log (\theta_0 [k])\bigl)
\end{align*}
}

Considering $\gpr$ as described in \Cref{sec:probmodel}, the above objective is solved by using Lagrange Multipliers, which is a strategy for finding the local maxima and minima of a function subject to equality constraints.
The multipliers $\lambda$ and $\mu$ are added to the constraints of $g$ and $h$ accordingly as follows:
\begin{align*}
    &g(p) = \sum_{w \in W(S[c])} p(w, c) = 1\\
    &h(q) = \sum_{w \in W(S[c])} q(w, c) = 1
\end{align*}

Hence, the objective function is now:
\begin{align*}
    &\argmax_{p,q}\Biggl(\Bigl(\sum_{w\in W(S[c])} A^+(w, c) \cdot \log (\alpha\cdot q(w, c) + (1-\alpha)\cdot p(w, c)) + A^-(w, c) \cdot \log p(w, c) \Bigl) \\
    &-\lambda(g(p) -1) -\mu(h(q) -1)\Biggl)
    = \underset{p, q}{\argmax} \, \Phi
\end{align*}
 Solving the optimization problem leads to the following equations:
\begin{align}
    &\dfrac{\partial \Phi}{\partial \lambda} = \sum_{w\in W(S[c])} {p(w, c)} -1 = 0 \label{eq:lambda_grad}\\ 
    &\dfrac{\partial \Phi}{\partial \mu} = \sum_{w\in W(S[c])} {q(w, c)} -1 = 0 \label{eq:mu_grad}\\
    &\forall w\in W(S[c]): \dfrac{\partial \Phi}{\partial {p(w, c)}} = \dfrac{(1-\alpha) \cdot A^+(w, c)}{\alpha\cdot q(w, c) + (1-\alpha)\cdot p(w, c)} + \dfrac{A^-(w, c)}{p(w, c)} -\lambda = 0 \label{eq:teta0_grad}\\
    &\forall w\in W(S[c]): \dfrac{\partial \Phi}{\partial {q(w, c)}} = \dfrac{\alpha \cdot A^+(w, c)}{\alpha\cdot q(w, c) + (1-\alpha)\cdot p(w, c)} -\mu = 0 \label{eq:teta1_grad}
\end{align}

from equations \Cref{eq:teta0_grad} and \Cref{eq:teta1_grad}:
\begin{align}
    &q(w, c) = \dfrac{(1-\alpha) \cdot (A^+(w, c) + A^-(w, c) + \lambda \cdot p(w, c)) \cdot p(w, c)}{\alpha\cdot (\lambda \cdot p(w, c) -  A^-(w, c))}\\
    &q(w, c) = \dfrac{\alpha \cdot A^+(w, c) - \mu \cdot (1 - \alpha) \cdot p(w, c)}{\alpha \cdot \mu} \label{eq:teta1}
\end{align}

comparing those equations results in:
\begin{equation}
     p(w, c) = \dfrac{\alpha \cdot A^-(w, c)}{\alpha \cdot \lambda - (1-\alpha)\cdot \mu} \label{eq:teta0}
\end{equation}

We substitute this value of $p(w, c)$ in \Cref{eq:lambda_grad}:
\begin{equation*}
     \sum_{w\in W(S[c])} \dfrac{\alpha \cdot A^-(w, c)}{\alpha \cdot \lambda - (1-\alpha)\cdot \mu} = 1
\end{equation*}

and get: 
\begin{equation}
     \lambda = \dfrac{(\sum_{w\in W(S[c])} (\alpha \cdot A^-(w, c))) + (1-\alpha)\cdot \mu}{\alpha} \label{eq:lambda}
\end{equation}

Substitute the value of $q(w, c)$ of \Cref{eq:teta1} in \Cref{eq:mu_grad}:
\begin{equation*}
     \sum_{w\in W(S[c])} \dfrac{\alpha \cdot A^+(w, c) - \mu \cdot (1 - \alpha) \cdot p(w, c)}{\alpha \cdot \mu} = 1
\end{equation*}

And using \Cref{eq:lambda_grad} which sums $p(w, c)$ to 1:
\begin{equation*}
     \mu = \alpha \cdot \sum_{w\in W(S[c])} A^+(w, c)
\end{equation*}

Substitute this value of $\mu$ in \Cref{eq:lambda}:
\begin{align*}
    \lambda = \sum_{w\in W(S[c])} A^-(w, c) + (1-\alpha)\sum_{w\in W(S[c])} A^+(w, c)
\end{align*}

$\lambda$ and $\mu$ resulted as positive values, and so the constraints hold. \\
Using those found values in \Cref{eq:teta0} yields:
\begin{align*}
    & p(w, c) = \dfrac{A^-(w, c)}{\sum_{w'\in W(S[c])}A^-(w', c)}\\
\end{align*}

And finally using all found values in \Cref{eq:teta1}:
\begin{align*}
    q(w, c) = \dfrac{\dfrac{A^+(w, c)}{\sum_{w'\in W(S[c])}A^+(w', c)} -(1-\alpha)\dfrac{A^-(w, c)}{\sum_{w'\in W(S[c])}A^-(w', c)}}{\alpha} 
\end{align*}

To sum up, the solution to those equations results in a single solution under the constraints and assumptions mentioned, which is a local maximum to the original optimization problem:
\begin{align*}
    &\lambda = \sum_{w\in W(S[c])} A^-(w, c) + (1-\alpha)\sum_{w\in W(S[c])} A^+(w, c)\\
    &\mu = \alpha \cdot \sum_{w\in W(S[c])} A^+(w, c)\\
    &\forall w\in W(S[c]): p(w, c) = \dfrac{A^-(w, c)}{\sum_{w'\in W(S[c])}A^-(w', c)}\\
    &\forall w\in W(S[c]): q(w, c) = \dfrac{\dfrac{A^+(w, c)}{\sum_{w'\in W(S[c])}A^+(w', c)} -(1-\alpha)\dfrac{A^-(w, c)}{\sum_{w'\in W(S[c])}A^-(w', c)}}{\alpha}
\end{align*}


\section{Additional Results}
\begin{figure}
    \includegraphics[width=1.0\linewidth]{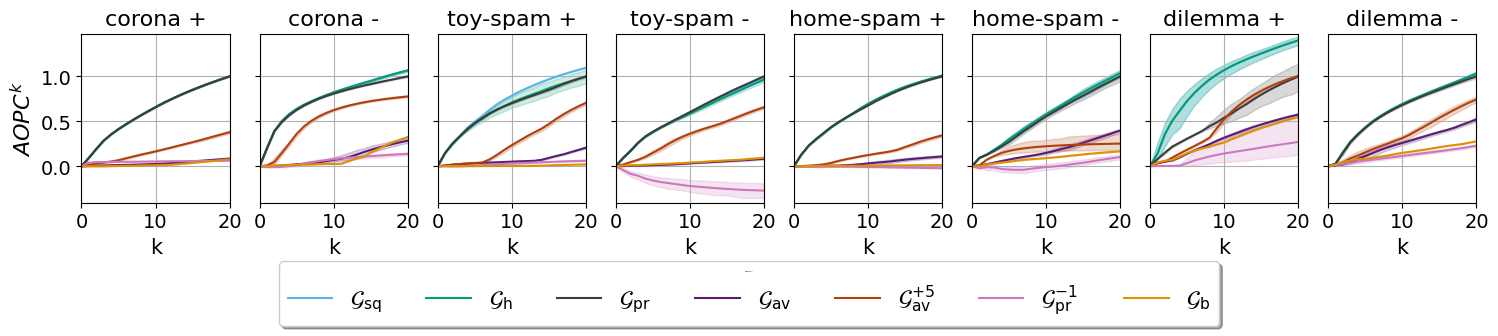}
    \caption{$\aopck(\G,c)$ of different aggregation functions $\G$ (y-axis) for varying $k$ (x-axis). Bert is used as the model.}
    \label{fig:Bert aggregation}
\end{figure}

\begin{figure}
    \includegraphics[width=1.0\linewidth]{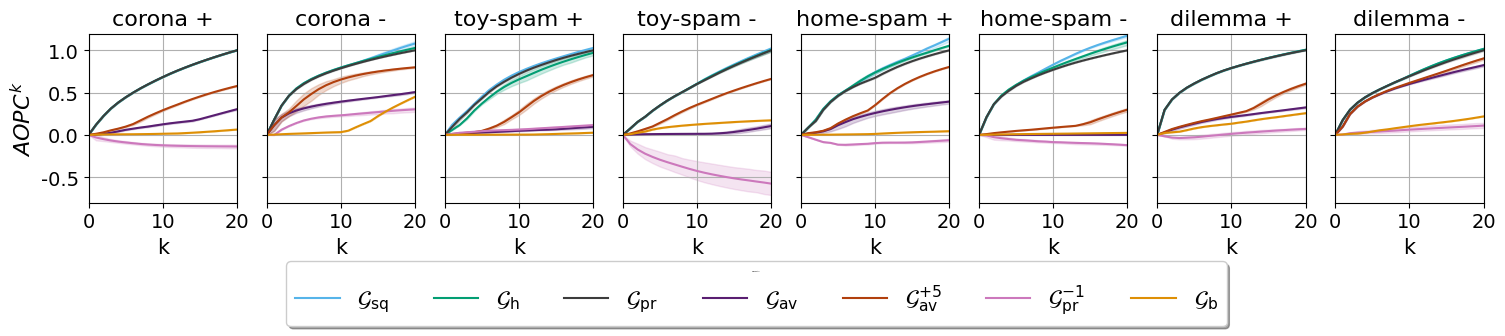}
    \caption{$\aopck(\G,c)$ of different aggregation functions $\G$ (y-axis) for varying $k$ (x-axis). Logistic Regression is used as the model.}
    \label{fig: Logistic aggregation}
\end{figure}

This section contains results of the experiments conducted in \Cref{subseq:results} for the Bert and Logistic Regression models. The following graphs present similar characteristics to those presented in \Cref{subseq:results} for Deberta. \Cref{fig:Bert aggregation} and \Cref{fig: Logistic aggregation} which compare aggregations both show $\gpr$ begins with a steep positive curve, and an overall height significantly higher than the baselines. $\gave$ is inferior to other functions as well, and $\gsqrt$ and $\gh$ perform similarly to $\gpr$.

\begin{figure}
    \includegraphics[width=1.0\linewidth]{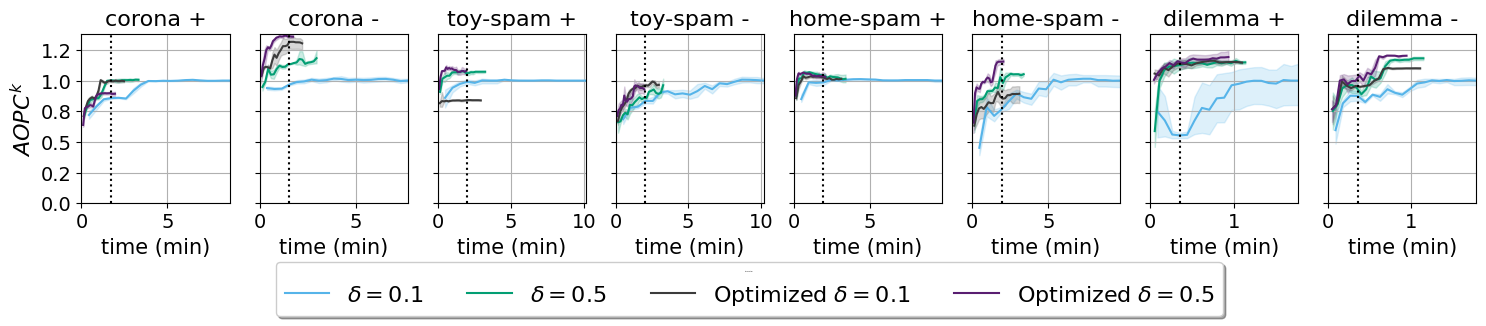}
    \caption{$\aopck(\G,c)$ for different versions $\G$ of $\gpr$ and $k=20$, as a function of the computation time. ``Optimized'' refers to the combination of all optimizations described in \Cref{sec:optimizations}. Bert is used as the model.}
    \label{fig:Bert Optimizations}
\end{figure}

\begin{figure}
    \includegraphics[width=1.0\linewidth]{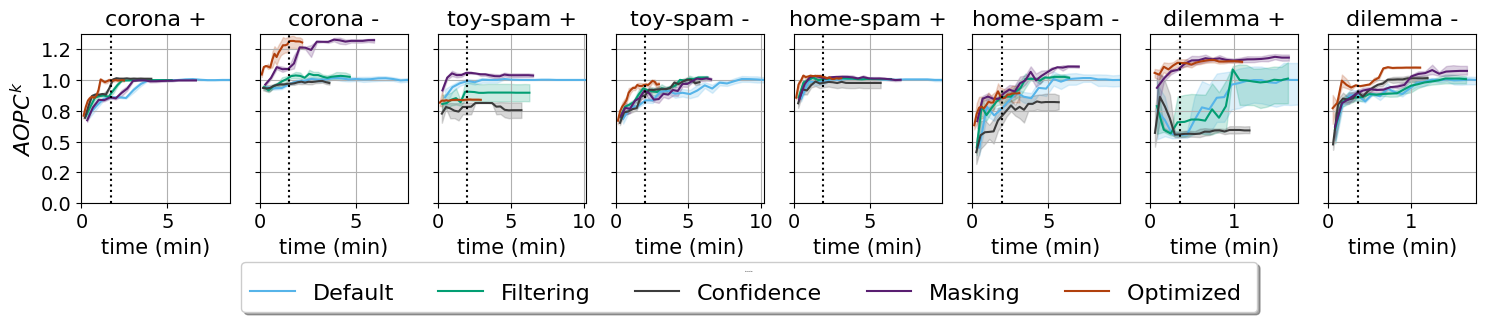}
    \caption{$\aopck(\G,c)$ for different optimizations of $\gpr$ and $k=20$, as a function of the computation time. ``Optimized'' refers to the combination of all optimizations. Bert is used as the model.}
    \label{fig:Bert Optimizations Sep}
\end{figure}

\begin{figure}
    \centering
    \includegraphics[width=1.0\linewidth]{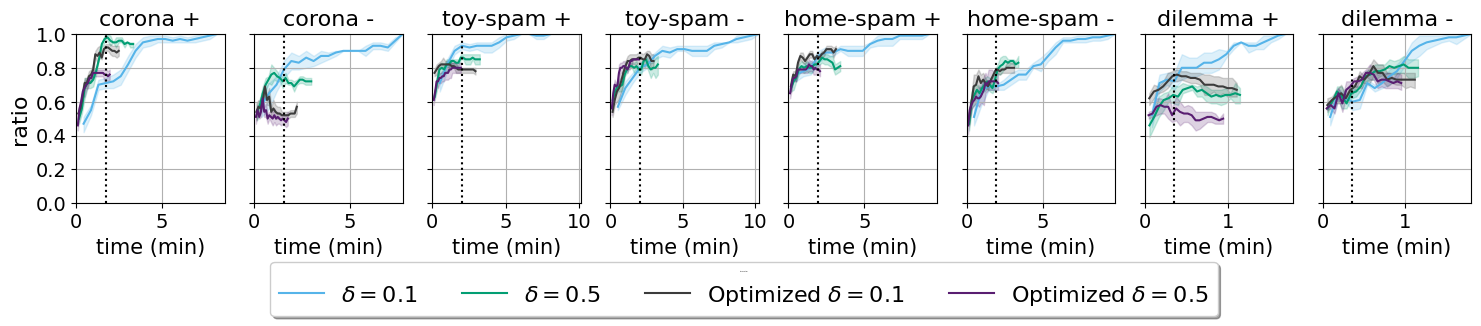}
    \caption{Ratio of shared terms for different versions of $\gpr$. Bert is used as the model.}
    \label{fig:Bert Optimizations percents}
\end{figure}

\begin{figure}
    \includegraphics[width=1.0\linewidth]{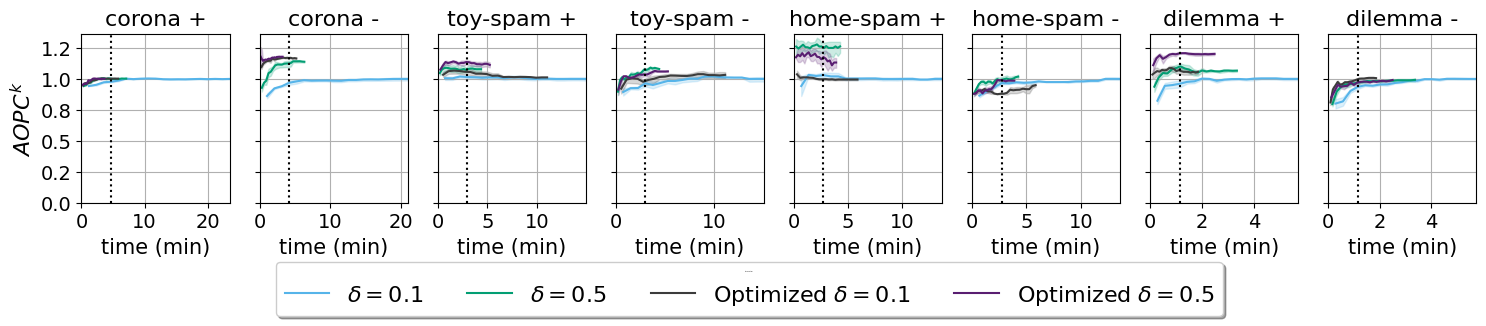}
    \caption{$\aopck(\G,c)$ for different versions $\G$ of $\gpr$ and $k=20$, as a function of the computation time. ``Optimized'' refers to the combination of all optimizations described in \Cref{sec:optimizations}. Logistic Regression is used as the model.}
    \label{fig:Logistic Optimizations}
\end{figure}

\begin{figure}
    \includegraphics[width=1.0\linewidth]{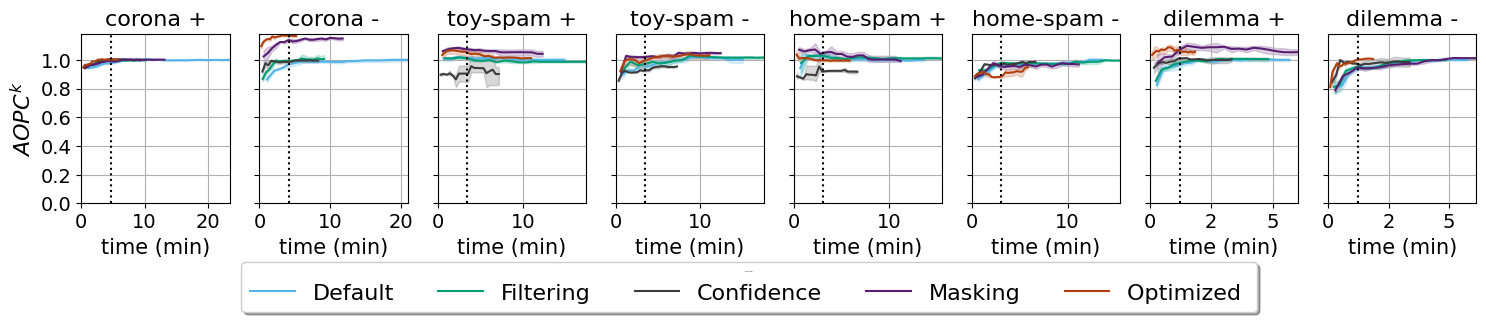}
    \caption{$\aopck(\G,c)$ for different optimizations of $\gpr$ and $k=20$, as a function of the computation time. ``Optimized'' refers to the combination of all optimizations. Logistic Regression is used as the model.}
    \label{fig:Logistic Optimizations sep}
\end{figure}

\begin{figure}
    \centering
    \includegraphics[width=1.0\linewidth]{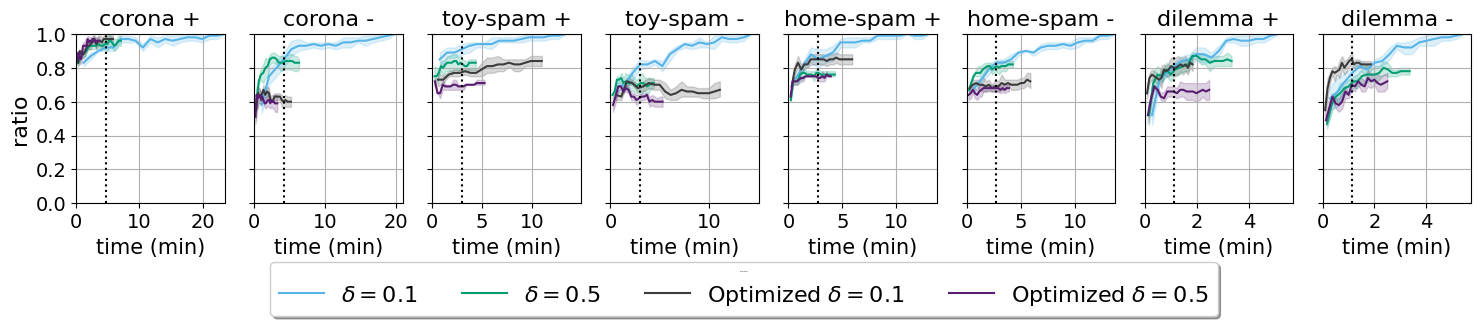}
    \caption{Ratio of shared terms for different versions of $\gpr$. Logistic Regression is used as the model.}
    \label{fig:Logistic optimizations percents}
\end{figure}

\Cref{fig:Bert Optimizations}, \Cref{fig:Bert Optimizations Sep}, \Cref{fig:Logistic Optimizations} and \Cref{fig:Logistic Optimizations sep} measure the quality of the set of $k$ ($=20$) candidates at different times during the computation of the algorithm until its completion. Similarly to their equivalent charts in \Cref{subseq:results}, \Cref{fig:Bert Optimizations} and \Cref{fig:Logistic Optimizations} show that increasing $\delta$ improves both the running time and quality of $\gpr$, without a drop in quality, and that combining the remaining optimizations shortens the overall running time. Moreover, applying each optimization individually (\Cref{fig:Bert Optimizations Sep},  \Cref{fig:Logistic Optimizations sep}) shortens the overall aggregation time. The \emph{confidence} and \emph{masking} optimizations also retain the property of being the worst and best optimizations in terms of quality accordingly, and applying all of the optimizations combined is the most efficient but costs in quality.
\Cref{fig:Bert Optimizations percents} and \Cref{fig:Logistic optimizations percents}  measure the percentage of shared terms between $T_c$ of $\gpr$ to that of its optimizations. While both are mostly similar to \Cref{fig:time percent optimizations}, we can see bigger drops in ratio in some cases, like in ``corona -''. However, this difference in the ratio of shared doesn't indicate a drop in quality necessarily, as seen in \Cref{fig:Bert Optimizations} and \Cref{fig:Logistic Optimizations}.

\end{document}